\documentclass[sigconf, table]{acmart}
\usepackage{booktabs} 
\usepackage{tabularx}
\usepackage{ amssymb }
\usepackage{ dsfont }

\acmDOI{10.475/123_4}

\acmConference[First Workshop On Machine Learning for Computer Systems]{ACM HPDC}{June 2018}{Tempe, AZ USA}
\acmYear{2018}
\copyrightyear{2018}

\hypersetup{draft}

\begin{document}
\title[Interpretable System Log Anomaly Detection]{
    Recurrent Neural Network Attention Mechanisms\\for Interpretable System Log Anomaly Detection}

\author{Andy Brown}
\authornote{First and second authors contributed equally to this work.}
\affiliation{%
  \institution{Western Washington University}
}
\email{browna52@wwu.edu}

\author{Aaron Tuor}
\authornotemark[1]
\affiliation{%
  \institution{Pacific Northwest National Laboratory}
}
\email{aaron.tuor@pnnl.gov}

\author{Brian Hutchinson}
\authornote{Joint appointment at PNNL and WWU (brian.hutchinson@pnnl.gov).}
\affiliation{%
  \institution{Western Washington University}
}
\email{brian.hutchinson@wwu.edu}

\author{Nicole Nichols}
\affiliation{%
  \institution{Pacific Northwest National Laboratory}
}
\email{nicole.nichols@pnnl.gov}
\renewcommand{\shortauthors}{A. Brown et al.}

\begin{abstract}
Deep learning has recently demonstrated state-of-the art performance on key tasks related to the maintenance of computer systems, such as  intrusion detection, denial of service attack detection, hardware and software system failures, and malware detection.  
In these contexts, model interpretability is vital for administrator and analyst to trust and act on the automated analysis of machine learning models. Deep learning methods have been criticized as black box oracles which allow limited insight into decision factors.
In this work we seek to ``bridge the gap'' between the impressive performance of deep learning models and the need for interpretable model introspection. 
To this end we present recurrent neural network (RNN) language models augmented with attention for anomaly detection in system logs. Our methods are generally applicable to any computer system and logging source.
 By incorporating attention variants into our RNN language models we create opportunities for model introspection and analysis without sacrificing state-of-the art performance.
 We demonstrate model performance and illustrate model interpretability on an intrusion detection task using the Los Alamos National Laboratory (LANL) cyber security dataset, reporting upward of 0.99 area under the receiver operator characteristic curve despite being trained only on a single day's worth of data.
\end{abstract}
%
%



\maketitle

\section{Introduction}
 System log analysis is critical for a wide range of tasks in maintaining large scale computer systems such as enterprise computer networks and high performance computing clusters. 
 These include security tasks such as intrusion detection, insider threat detection, and malware detection, as well as more general maintenance tasks such as detecting hardware failure and modeling data or traffic flow patterns. 
 Extracting knowledge from information rich system logs is complicated by several factors:
 \begin{enumerate}
 \item Log sources can generate terabytes of data per day.
\item Labeled data for application areas of interest is often scarce, unbalanced, or system specific.              
\item Actionable information may be obscured by complex, undiscovered relationships across logging sources and system entities (e.g. users, PCs, processes, nodes).
\end{enumerate}

Due to these factors, unaided human monitoring and assessment is impractical, so considerable research has been directed to automated methods for visualization and analysis of system logs. 
Furthermore, as administrative decisions may be of considerable consequence to organizations and associated persons, it is crucial to have some understanding of the factors involved in automated decision processes, even for highly effective algorithms. 
  
  Addressing these factors, we present unsupervised recurrent neural network (RNN) language models for system log anomaly detection. 
  By modeling the normal distribution of events in system logs, the anomaly detection approach can discover complex relationships buried in these logs. 
  Since the methods are unsupervised, the models do not depend on the time consuming and otherwise expensive procurement of labeled data. 
  Our language modeling framework requires little to no feature engineering: it is applicable to any serializable logging source. 
  Further, the models are trained online using bounded resources dictated by the daily volume of the log sources.
  
   Our main contributions in this work are twofold: 1) we evaluate the effectiveness of augmenting RNN language models with several attention mechanisms specifically designed for system log anomaly detection, and 2) we illustrate how model introspection in these systems is made possible by the attention mechanisms. 
  
  \section{Related Work}
Recently, several researchers have used Long Short-Term Memory (LSTM) Networks \cite{Hochreiter1997} in system log analysis.
\citet{zhang2016automated} use clustering techniques on the raw text from multiple log sources to generate feature sequences fed to an LSTM for hardware and software failure predictions. 
 \citet{du2017deeplog} employ customized parsing methods on the raw text of system logs to generate sequences for LSTM Denial of Service attack detection. In contrast to these methods our approach works directly with raw text with no pre-processing beyond tokenization using known field delimiters. 
 Others have incorporated LSTM networks to preprocess sequences of process API calls as components to malware detection systems \cite{pascanu2015malware} trained on labeled malware examples. 
 
Attention-equipped LSTM models have been used to improve performance on complex sequence modeling tasks. 
Attention provides a dynamic weighted average of values from different points in a calculation during the processing of a sequence to provide long term context for downstream discriminative or generative prediction. 
In recent work \cite{salton2017attentive, daniluk2017frustratingly, yogatama2018memory}, researchers have augmented LSTM language models with attention mechanisms in order to add capacity for modeling long term syntactic dependencies. 
\citet{yogatama2018memory} characterize attention as a differentiable random access memory. They compare attention language models with differentiable stack based memory \cite{grefenstette2015learning} (which provides a bias for hierarchical structure), 
demonstrating the superiority of stack based memory on a verb agreement task with multiple attractors. 
\citet{daniluk2017frustratingly} explore three additive attention \cite{bahdanau2014neural} mechanisms  with successive partitioning of the output of the LSTM; splitting the output into separate key, value, and prediction vectors performed best, likely due to removing the need for a single vector to encode information for multiple steps in the computation. In contrast we augment our language models with dot product attention \cite{luong2015effective, vaswani2017attention}, but also use separate vectors for the components of our attention mechanisms. 

Many decision processes raise ethical dilemmas \cite{mittelstadt2016ethics} or are applied in critical domains with high consequence. 
Such factors necessitate human interpretation of how a model is generating its predictions to ensure acceptable results. 
\citet{vellido2012making} observe the gap between data modeling, knowledge extraction, and potential machine learning solutions, underscoring the need for interpretable automated decision processes. 
However, interpretability has multiple goals that are not always aligned with production of the most generalizable model architecture \cite{lipton2016mythos}.
 Hence, there is currently a large research focus on making interpretable deep learning algorithms for sensitive and critical application areas.  
 Some proposed model introspection techniques include dimensionality reduction \cite{wold1987principal}, 
 analysis of intermediate layers \cite{alain2016understanding} and saliency based methods \cite{changinterpreting,nagasubramanianexplaining}.
 In contrast to other deep learning components, attention mechanisms allow an immediate view into what factors are affecting model decisions. 
 \citet{xu2015show} examine attention weights to determine what convolutional neural networks are ``looking'' at while making predictions. 
 Similarly, \citet{rocktaschel2015reasoning} analyze matrices of word-to-word attention weights for insight into how their LSTM entailment classifier reasons about sentences. 
 We apply the same concept to explore what factors our models attend over when predicting anomaly scores.

\section{Methods}
\begin{figure}
	\includegraphics[width=.47\textwidth]{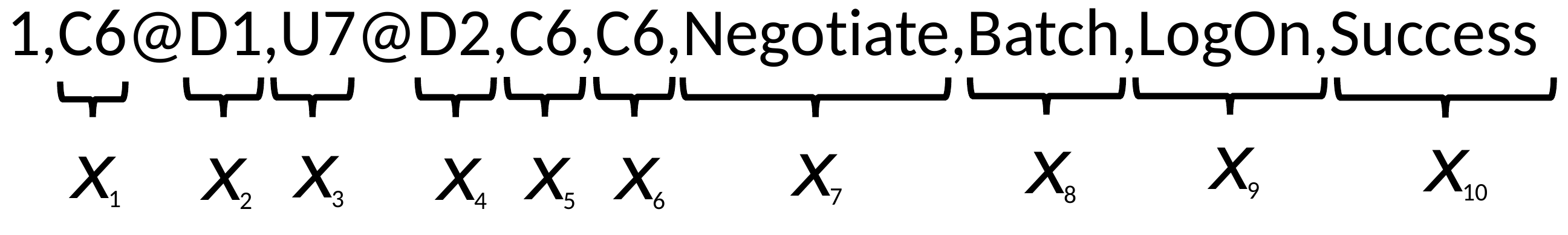}
	\includegraphics[width=.47\textwidth]{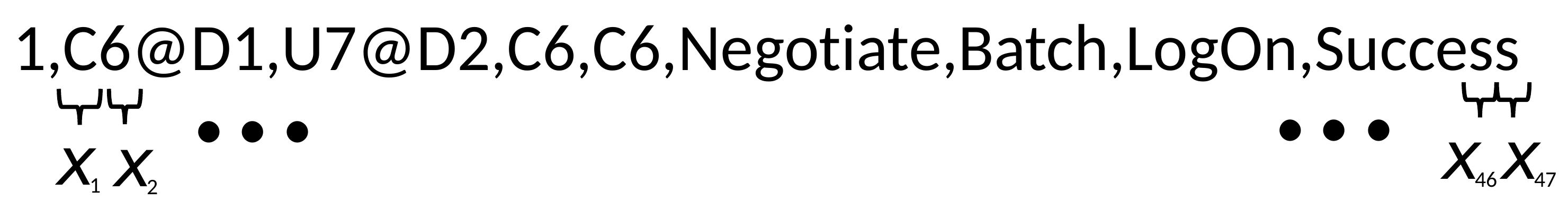}
	\caption{Top: Word tokens; Bottom: Character tokens} \label{fig:token}
\end{figure}
Here we describe the unsupervised language modeling framework and its extension via five variations of attention. In each case, the language models consume a sequence of log-line tokens and output log-line-level anomaly scores. 

\subsection{Preliminaries}
\subsubsection{Language Modeling}
We assume that each log-line consists of a sequence of $T$ tokens: $x_{(1:T)} = x_{(1)}, x_{(2)}, \dots, x_{(T)}$.  Each token $x_{(t)} \in \mathds{V}$, where $\mathds{V}$ denotes the vocabulary.  A {\it language model} is a model that assigns probabilities to sequences: $P(x_{(1:T)})$.  A language model often evaluates the probability of a sequence using the chain rule of probability:
\begin{equation}
    P(x_{(1:T)}) = \prod_{t=1}^T P(x_{(t)}|x_{(<t)})
\end{equation}
where $x_{(<t)}$ denotes the (potentially empty) sequence of tokens from $x_{(1)}$ to $x_{(t-1)}$.
The conditional probabilities on the righthand side can be modeled with a recurrent neural network, as will be described in the Section \ref{subsec:cyberlm}.

Our data consist of a series of log-lines, each affiliated with a user.  We denote user $u$'s $i$th log-line with $x^{(u,i)}_{(1:T)}$, but omit the superscript when it is non-essential. 
Our language models all output a single anomaly score, the negative log-likelihood, for each log-line.  

\subsubsection{Tokenization}

Figure \ref{fig:token} illustrates two methods to partition log lines into sequences of tokens: word and character tokenization. For word based language modeling, the tokens are the fields of the CSV format log file. 
The user fields are further split on the ``@'' character to generate user name and domain tokens. A frequency threshold is applied to replace infrequent words with an ``out of vocabulary'' (OOV) token; a value must occur in a field at least 40 times to be added to the vocabulary. The OOV token ensures that our models will have non-zero probabilities when encountering previously unseen words during evaluation.

For character based language modeling we use a primitive vocabulary consisting of printable ASCII characters. This circumvents the out of vocabulary issues with the word model. Delimiters are left in the character inputs to give context of switching fields to the models.
For both word and character tokenization, the time field is ignored and not tokenized. 

\subsection[Cyber Anomaly Language Models]{Cyber Anomaly Language Models} \label{subsec:cyberlm}

We recently introduced a language modeling framework
for cyber anomaly \cite{tuor2017recurrent} that forms the starting point of this work.
The first of four models presented in \cite{tuor2017recurrent} is the ``Event Model'' (EM), which applies a standard LSTM \cite{Hochreiter1997} to the token sequences of individual events (log-lines).  In order to feed the categorical tokens $x_{(1:T)}$ into the model, we first perform an embedding lookup on each token to yield the sequence ${\bf x}_{(1:T)}$ (bold font), where each ${\bf x}_{(t)} \in \mathds{R}^{L_{emb}}$ for some embedding dimension hyperparameter, $L_{emb}$.  There are unique embedding vectors for each element in the vocabulary; these embedding vectors are parameters of the model, learned jointly with all other model parameters.  An LSTM maps an embedding vector sequence to a sequence of hidden vectors ${\bf h}_{(1:T)}$:%
\footnote{In this paper we assume all vectors are row vectors and adopt the notation convention of left multiplying matrices with row vectors (omitting the conventional transpose to avoid clutter). 
}
\begin{equation}
    \textrm{LSTM}(\textbf{x}_{(1:T)}) = {\bf h}_{(1:T)}
\end{equation}
Intuitively, ${\bf h}_{(t)}$ is a summary of the input sequence ${\bf x}_{(1:t)}$ defined by the same, standard LSTM equations used in \cite{tuor2017recurrent}.
%
%
Given the previous hidden state $\textbf{h}_{(t-1)}$, weight matrix $\textbf{W}$ and bias vector $\textbf{b}$, the probability distribution over the token at step $t$ is:
\begin{equation}
    \textbf{p}_{(t)} = \textrm{softmax} 
    \biggl(\underbrace{\mathbf{h}_{(t-1)}}_{ L_h} 
    \underbrace{\vphantom{\mathbf{h}_{(t-1)}} \mathbf{W}}_{L_h \times |\mathds{V}|}+
    \underbrace{\vphantom{\mathbf{h}_{(t-1)}} \mathbf{b}}_{|\mathds{V}|}\biggr) 
    \in \mathds{R}^{|\mathds{V}|} \label{eqn:em_prob}
\end{equation}
This conditions each prediction on all tokens that precede it in the log-line.
The second model in \cite{tuor2017recurrent} is the Bidirectional Event Model (BEM), which updates Eqn. \ref{eqn:em_prob} to also incorporate the hidden state from a backward-running LSTM, with hidden vector $\textbf{h}_{(t+1)}^b$ and additional weight matrix $\textbf{W}^b$ as follows:
\begin{equation}
    \textbf{p}_{(t)} = \textrm{softmax} \left(\mathbf{h}_{(t-1)}\mathbf{W}+\mathbf{h}_{(t+1)}^b\mathbf{W}^b+\mathbf{b}\right)  
\end{equation}
The BEM conditions each prediction on the all of the other tokens in the log-line (preceding or following), for richer context.

\begin{figure}
    \includegraphics[width=.5\textwidth]{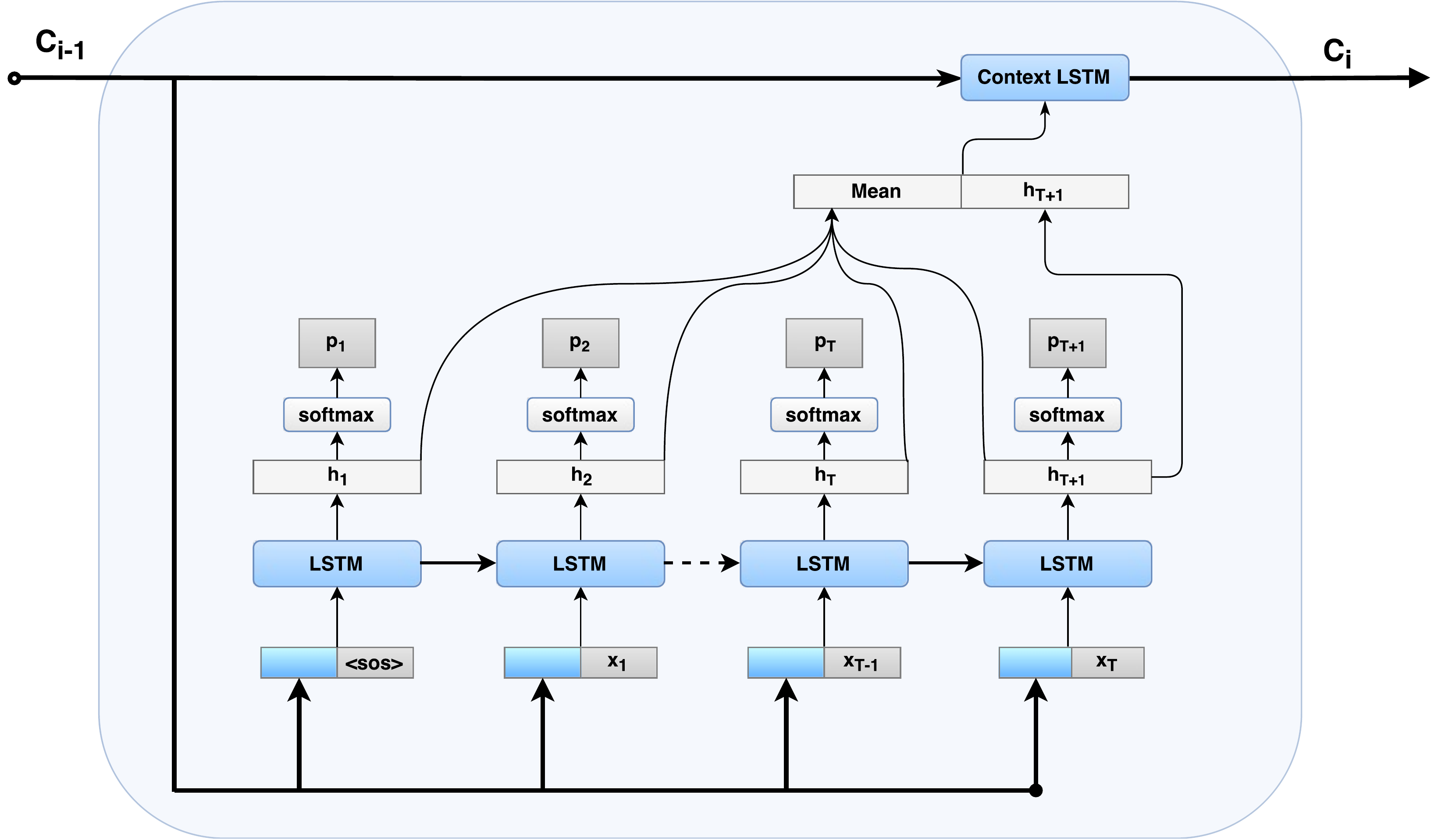}
    \caption{Tiered language model (T-EM) \cite{tuor2017recurrent}.} \label{fig:tiered} 
\end{figure}

The EM and BEM only condition predictions on other tokens in the same log line.  However, \citet{tuor2017recurrent} also introduce tiered language model variants that employ an ``upper tier'' LSTM to model a user's sequence of log-lines (see Fig. \ref{fig:tiered}).  
Each log-line is still modeled by an EM or BEM, but the input is the concatenation of embedding vectors ${\bf x}_t$ along with a {\it context vector} produced by the upper tier LSTM. 
The upper tier LSTM takes as input a summary of the lower-tier hidden states (the average lower-tier hidden state concatenated with the final hidden state).
The upper and lower tiers are trained jointly. For later reference, we name these models T-EM and T-BEM, respectively.

For all language models (including the tiered models which incorporate inter-log-line context) we optimize the model parameters by minimizing the negative log-likelihood produced by EM or BEM predictions. 
The negative log-likelihood minimization objective also serves as the anomaly score for the log line (less probable events receiving higher anomaly scores). 

\subsection{Attention} \label{subsec:attention}
\begin{figure}
	\includegraphics[width=.45\textwidth]{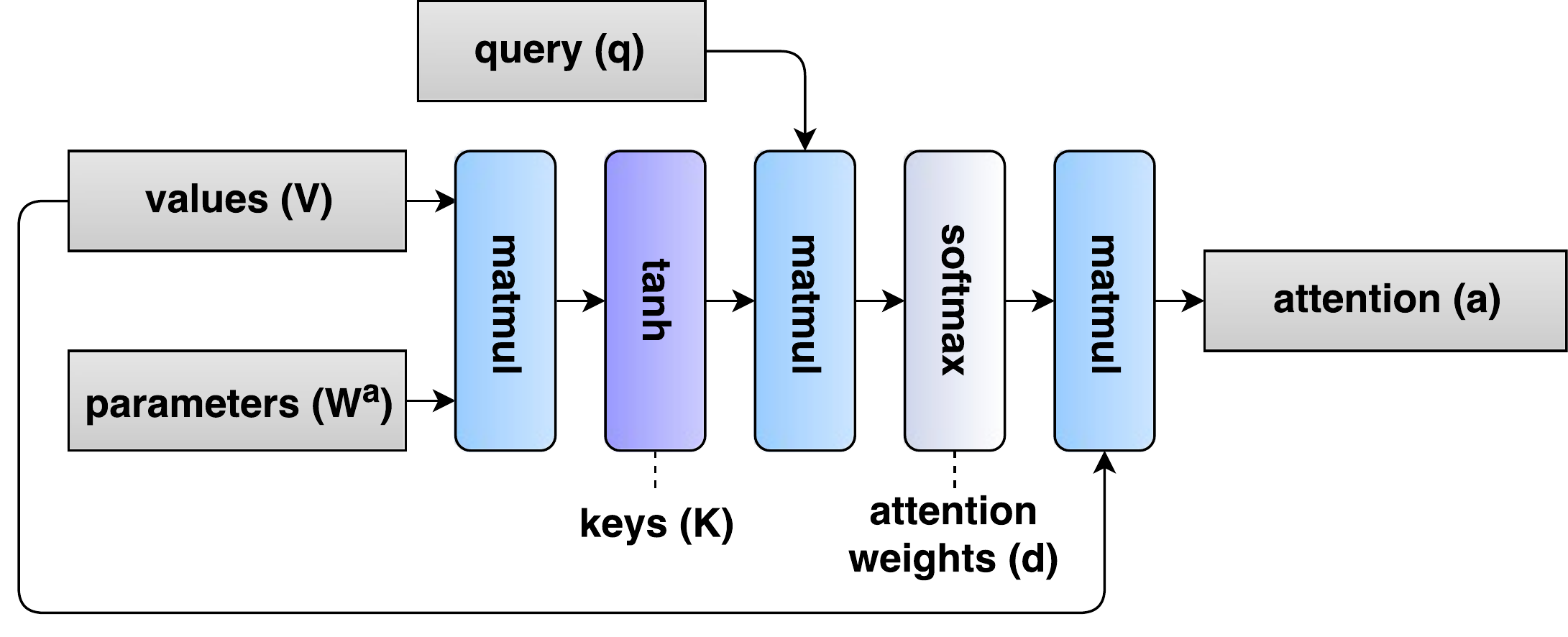}
	\caption{Dot Product Attention.} \label{fig:att}
\end{figure}
In this work we use dot product attention (Figure \ref{fig:att}), wherein an ``attention vector'' ${\bf a}$ is generated from three values: 1) a key matrix ${\bf K}$, 2) a value matrix ${\bf V}$, and 3) a query vector ${\bf q}$.
In this formulation, keys are a function of the value matrix: 
\begin{equation}
    \textbf{K} = \tanh(\textbf{V}\textbf{W}^a ),
\end{equation}
parameterized by $\textbf{W}^a$.
The importance of each timestep is determined by the magnitude of the dot product of each key vector with the query vector ${\bf q} \in \mathds{R}^{L_a}$ for some attention dimension hyperparameter, $L_a$. These magnitudes determine the weights, ${\bf d}$ on the weighted sum of value vectors, ${\bf a}$:
\begin{eqnarray}
    \textbf{d}     & = & \textrm{softmax}(\textbf{q}\textbf{K}^T) \\ 
    \textbf{a} & = & \textbf{d} \textbf{V} 
\end{eqnarray}

In an LSTM, the information relevant to a given prediction (e.g. of the next token, $x_{(t+1)}$) is accumulated and propagated via the LSTM's cell state, ${\bf c}_{(t)}$.  
For any given prediction, however, certain tokens are likely to be more relevant than others.  
Attention provides a mechanism for predictions to be directly, selectively conditioned on a subset of the relevant tokens.  
In practice, this is accomplished by making ${\bf p}_{(t)}$ a function of the concatenation of ${\bf h}_{(t-1)}$ with an attention vector  ${\bf a}_{(t-1)}$ that is a weighted sum over hidden states :

\begin{equation}
    \textbf{p}_{(t)} = \textrm{softmax} \left(
    \begin{bmatrix}
     {\bf h}_{(t-1)}& {\bf a}_{(t-1)}
     \end{bmatrix}
    \mathbf{W}+ \mathbf{b}\right) 
\end{equation}

This attention mechanism not only introduces shortcuts in the flow of information over time, allowing the model to more readily access the relevant information for any given prediction, but the weights on the weighted sum also yield insights into the model's decision process, aiding interpretability.

We first examine the case of adding attention to the standard EM.
Each token-step $t$ is associated with its own value matrix ${\bf V}_{(t)}$, and query vector ${\bf q}_{(t)}$.
The value matrix $\textbf{V}_{(t)}$ is the matrix of hidden states up to but excluding token-step $t$, where $L_h$ is the dimension of the LSTM hidden states.  These are the values over which the weighted sum will be performed.

\begin{equation}
    \textbf{V}_{(t)} = 
\begin{bmatrix}
\textbf{h}_{(1)} \\
\vdots \\
\textbf{h}_{(t-1)}
\end{bmatrix}  \in \mathds{R}^{(t-1) \times L_h}
\label{eqn:valuemat}
\end{equation}
From the value matrix and weight matrix $\textbf{W}^a \in \mathds{R}^{L_h \times L_a}$, we compute a set of keys for each token/step:
\begin{equation}
    \textbf{K}_{(t)} = \tanh(\textbf{V}_{(t)}\textbf{W}^a) \in \mathds{R}^{(t-1) \times L_a} 
\end{equation}
Then,
\begin{eqnarray}
    {\bf d}_{(t)} & = & \textrm{softmax}(\textbf{q}_{(t)} {\bf K}_{(t)}^T) \\
    {\bf a}_{(t)} & = & {\bf d}_{(t)} \textbf{V}_{(t)}
\end{eqnarray}

Our EM attention variants differ primarily in the definition of the query vector ${\bf q}_{(t)}$.
\subsubsection{Fixed Attention}
In the fixed variation of attention \cite{salton2017attentive} we let $\textbf{q}_{(t)} = {\bf q}$ for some fixed, learned vector ${\bf q}$ that is shared across all tokens/steps. This assumes some positions in the sequence are more important than others, but that importance does not depend on the token one is trying to predict.

\subsubsection{Syntax Attention}
Syntax attention differs from fixed attention in that $\textbf{q}_{(t)}$ is not shared across $t$.  This assumes that some tokens are more important than others and this importance depends on the position in the sequence for the token to predict, but not on the actual values for tokens $x_1,\dots,x_{t-1}$.

\subsubsection{Semantic Attention 1}
For the first ``semantic'' variation, our query is a a function of the current hidden state  and parameter matrix $\textbf{W}^{sem1} \in \mathds{R}^{L_h \times L_a}$: 
\begin{equation}
    \textbf{q}_{(t)} = \tanh(\textbf{h}_{(t)}\textbf{W}^{sem1})
\end{equation}

\subsubsection{Semantic Attention 2}
Instead of making ${\bf q}_{(t)}$ a function of ${\bf h}_{(t)}$, in this variant we interpret each ${\bf h}_{(t)}$ emitted from the LSTM as the concatenation of two vectors: ${\bf h}'_{(t)}$ and ${\bf q}_{(t)}$.  The query portion, ${\bf q}_{(t)}$ is used as before, but now the value ${\bf V}_{(t)}$ defined in Eqn. \ref{eqn:valuemat} contains ${\bf h}'_{(1)}$ through ${\bf h}'_{(t-1)}$.  Note that, per the LSTM equations, both ${\bf h}'_{(t)}$ and ${\bf q}_{(t)}$ will be fed back into the LSTM at time $t+1$.

\subsubsection{Tiered Attention}
As shown in Fig. \ref{fig:tiered}, in original formulation of the tiered model, the lower tier LSTM hidden states are averaged in the process of passing information from the lower tier to the upper tier.  Implementation of attention for the tiered language models replaces this mean with a weighted average via attention. Let $\textbf{V}^{(u,i)}$ be the lower tier hidden states for user $u$'s $i$th log line:

\begin{equation}
\textbf{V}^{(u,i)} = 
\begin{bmatrix}
    \textbf{h}_{(1)}^{(u,i)} \\
    \vdots \\
    \textbf{h}_{(T)}^{(u,i)}
\end{bmatrix}
\end{equation}

\begin{figure}
	\includegraphics[width=.5\textwidth]{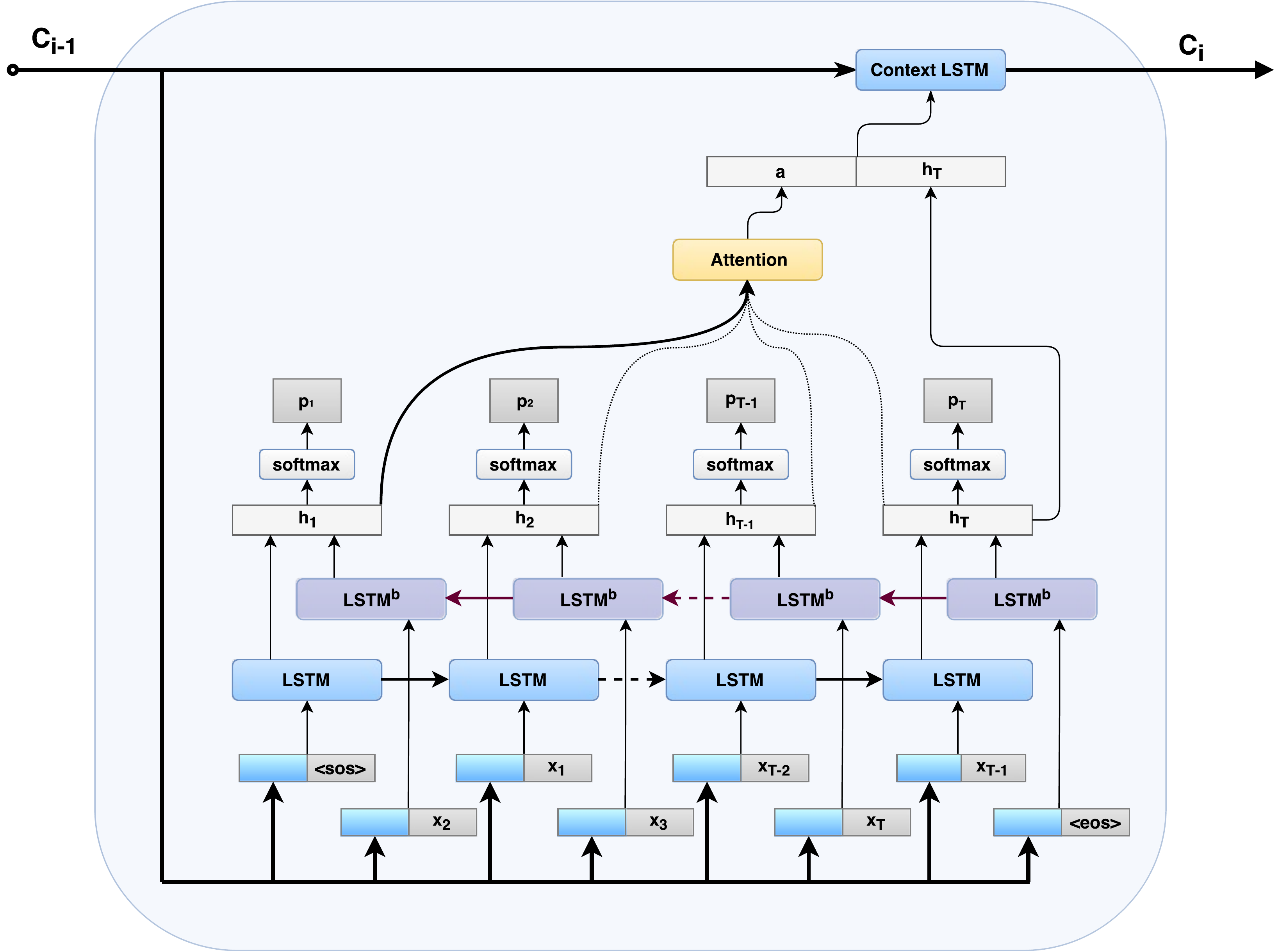}
	\caption{Tiered attention with bidirectional lower tier} \label{fig:btier}
\end{figure}

Let $\textbf{W}^{tier} \in \mathds{R}^{L_k \times L_a}$ and  $\textbf{W}^a  \in \mathds{R}^{L_a \times L_k}$ be parameter matrices. We then define the following attention mechanisms:
\begin{eqnarray}
    \textbf{q} & = & \tanh(\textbf{h}_{(T)}\textbf{W}^{(tier)})\\
    \textbf{K} & = & \tanh(\textbf{V}\textbf{W}^a )\\
    \textbf{d} & = & \textrm{softmax}(\textbf{q}\textbf{K}^{T}) \\ %
    \textbf{a} & = & \textbf{d} \textbf{V} %
\end{eqnarray}
We then replace the average of the hidden states in the tiered model with $\textbf{a}$. Note that each sequence shares weights $\textbf{W}^a$ and $\textbf{W}^{tier}$. The BEM tiered attention model (TA-BEM) is depicted in Figure \ref{fig:btier}.
\begin{figure}
	\includegraphics[width=.45\textwidth]{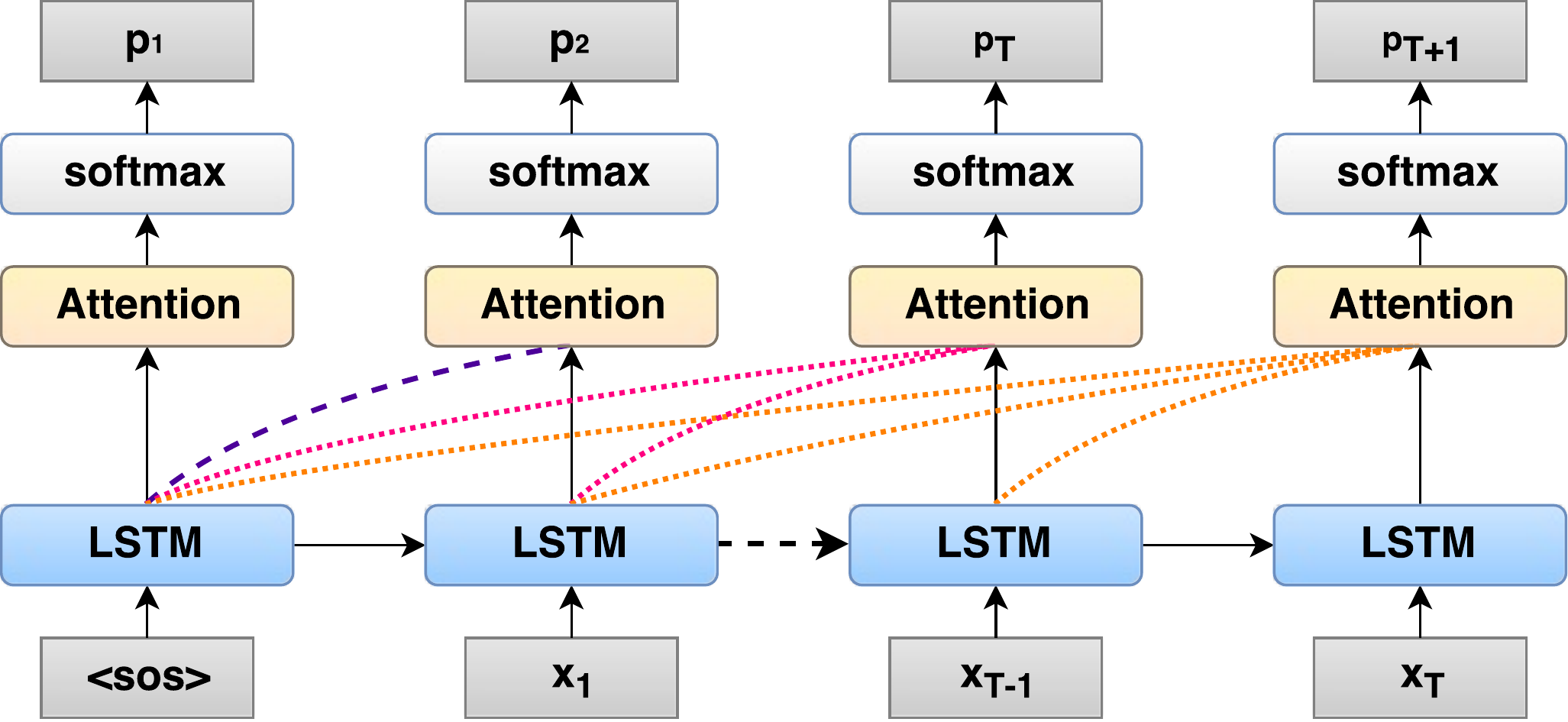}
	\caption{Event Model (EM) with attention. Dotted lines indicate which hidden states are being attended over.}
\end{figure}

\subsection{Online Training}
 We employ a syncopated online training algorithm, which both allows our model to continually adapt to changing distributions of activities on a network and to be deployed on high throughput streaming data sources. At the beginning of each day/cycle, the parameters of the current model are fixed for evaluation, thereby avoiding evolving anomaly score scale issues that could result from continuous online training. After anomaly scores have been calculated for the day's events we train on the current day's events. The days events are then discarded, bounding the storage demands of the algorithm to a day's worth of activity (plus model parameters). On the first day we do not evaluate as the model has not had a training phase yet.
 
 At the cost of the additional space complexity of storing two copies of the model parameters, the training and evaluation phases can be run concurrently. The evaluation and training parameters are then synced daily so that the evaluation copy is updated with the parameters of the training copy at the beginning of each day.

\section{Experiments}
This section discusses the data, experimental setup, evaluation metrics, and results assessing performance for the proposed methods.

\subsection{Data}
We evaluate our models on the publicly available LANL \cite{kent2016cyber} dataset. LANL consists of over one billion log lines collected over 58 consecutive days. The logs contain anonymized process, network flow, DNS, and authentication information. Interleaved are attacks by a red team. 
Our experiments focus on modeling the authentication logs, which contain the following fields:
\begin{quote}
\small
 Source user, Destination user, Source pc, Destination pc, Authentication type, Logon type, Authentication orientation,  Success/failure.
 \end{quote}
These events are collected from desktop PCs, servers, and active directory servers using the Windows OS. We filter automated system events by discarding all log-lines that have a machine listed as the source user. Red team event log-lines are indicated in the dataset. As our models are fully unsupervised, we use the red team labels only for evaluation of model performance. 

\subsection{Experimental Setup}
To assess our model's ability to spin up rapidly and detect anomalies with minimal burn-in time, we limit our scope to days 7 and 8, which contain 1 and 261 red team events respectively. Each of these days contains over seven million user log lines. 
We chose these particular days for evaluation because day 8 has the largest number of red events in the dataset. 
The entire experimental process is therefore 1) train on day 7, 2) evaluate on day 8.
Further simulating a rapid deployment process, we performed no hyper-parameter tuning.
Our learning rate is fixed to 0.01; 
we train using the ADAM \cite{DBLP:journals/corr/KingmaB14} optimizer; 
the minibatch size is 64; our LSTMs have a single layer with 128 hidden units; our token embedding size is 128 and our attention size is 128. 
To estimate model variability, we trained each model five times with the fixed hyper-parameters but different random weight initializations. 
In our results section we report statistics over the five runs.

\subsection{Metrics and Score Normalization}
We evaluate our results using the area under the receiver operating characteristic curve (AUC ROC).
ROC plots the true positive rate against the false positive rate as the detection threshold is swept.
Perfect detection yields an AUC of 1 and random guessing yields 0.5.
Recall that our anomaly scores, $z^{(u,i)}$ are given by the sum of the negative log probabilities of the tokens in line ${\bf x}^{(u,i)}_{(1:T)}$.
For word tokenization, we center each user's anomaly score:
\begin{equation}
    z^{(u,i)} \leftarrow z^{(u,i)} - \frac{1}{N_u} \sum_i z^{(u,i)}, \forall_u,
\end{equation}
where $N_u$ is the number of events by user $u$ in the day.  This reduces inter-user anomaly bias, which can stem from the uneven distribution of user name tokens.
 This normalization is unnecessary for the character tokenization, as the user names are composed from a common character vocabulary. 

\subsection{Results}
\begin{table}
	\centering
	\begin{tabularx}{.78\linewidth}{l  c  c  c  c  }

		{\bf Model}&{\bf Mean}&{\bf Max}&{\bf Min} &{\bf Std. Dev.} \\
		\midrule
		EM&0.968 &0.976&0.964&0.005\\
		\rowcolor[gray]{0.9}BEM&0.976 &0.981&0.972&0.003\\
		&&&&\\
		\multicolumn{5}{c}{\bf EM with attention}\\
		Fixed&0.974&0.976&0.972&0.001\\
		\rowcolor[gray]{0.9}Syntactic&0.972&0.975&0.967&0.004\\
		Semantic 1&0.975&0.980&0.971&0.004\\
		\rowcolor[gray]{0.9}Semantic 2&0.973&0.976&0.968&0.003\\
		&&&&\\
		\multicolumn{5}{c}{\bf Tiered LSTM variants}\\
		T-EM&0.984&0.989&0.977&0.005\\
		\rowcolor[gray]{0.9}T-BEM&0.987&0.989&0.985&0.002\\
		TA-EM&0.985&0.991&0.979&0.004\\
		\rowcolor[gray]{0.9}TA-BEM&0.988&0.991&0.984&0.003\\
	\end{tabularx}
	\caption{AUC statistics for word tokenization models} \label{tab:wordresults}
\end{table}
\begin{table}
	\centering
	\begin{tabularx}{.78\linewidth}{l  c  c  c  c  }
		{\bf Model}&{\bf Mean}&{\bf Max}&{\bf Min} &{\bf Std. Dev.} \\
		\midrule
		EM&0.965&0.969&0.961&0.003\\
		\rowcolor[gray]{0.9}BEM&0.985&0.987&0.982&0.002\\
		&&&&\\
		\multicolumn{5}{c}{\bf EM with attention}\\
		Fixed&0.963&0.971&0.937&0.015\\
		\rowcolor[gray]{0.9}Syntactic&0.967&0.973&0.963&0.004\\
		Semantic&0.975&0.977&0.971&0.003\\
		\rowcolor[gray]{0.9}Semantic 2&0.972&0.977&0.967&0.004\\
		&&&&\\
		\multicolumn{5}{c}{\bf Tiered LSTM variants}\\
		T-EM&0.977&0.988&0.967&0.008\\
		\rowcolor[gray]{0.9}T-BEM&0.992&0.992&0.991&0.000\\
		TA-EM&0.982&0.984&0.979&0.002\\
		\rowcolor[gray]{0.9}TA-BEM&0.991&0.992&0.990&0.001\\
	\end{tabularx}
	\caption{AUC statistics for character tokenization models} \label{tab:charresults}
\end{table}

In this section we discuss performance of the different attention mechanisms. 
We note that variance of model performance across random parameter initializations is quite low for most models. 
This low variance given only a single day of pretraining suggests our method behaves predictably despite rapid deployment. 

\subsubsection{Word Tokenization Models}
Table \ref{tab:wordresults} shows AUC statistics for the word tokenization model experiments.
Comparing the word level LSTM baselines, the BEM outperforms the EM.
However, adding attention to the EM improves performance to match the BEM. 
All variations of attention have very similar AUC scores. 
We hypothesize that the word model equally benefits from Syntax and Semantic attention, as it has a consistent syntax structure. 
 Tiered word models with attention do not demonstrate as significant performance gains, however, both forward and bidirectional attention models trend slightly upwards in mean and maximum values from their non-attention counterparts. 

\subsubsection{Character Tokenization Models}
As shown in Table \ref{tab:charresults}, 
the Fixed and Syntax attention models appear ill-suited for character-based models with variable length fields; neither Fixed nor Syntax attention improve performance here, and the character EM model augmented with Fixed attention has a standard deviation 2-15 times that of other models. 
In contrast, semantic variants, where the attention weights are a function of the current input as opposed to sequence position, do improve performance but are not on par with the BEM. 
For the tiered models, we see little difference by incorporating attention, suggesting the shortcuts introduced by attention are unnecessary to propagate user context across log-lines.
One interesting outcome is that a tiered model with either attention or a bidirectional lower tier has reduced variance across random initializations by a large factor for the character models. 

\section{Analysis}
While attention performs comparably to bidirectionality, it offers substantial advantages in its interpretability.  
Investigating which fields the model is attending to (and when) offers clues to its decision-making.
In this section we illustrate two approaches to analysis of attention-equipped LSTM language models: 1) Analysis of global model behavior from summary statistics of attention weights, and 2) analysis of particular model decisions from case studies of attention weights and language model predictions. 

\subsection{Global Behavior}
\begin{figure}
	\includegraphics[width=.35\textwidth]{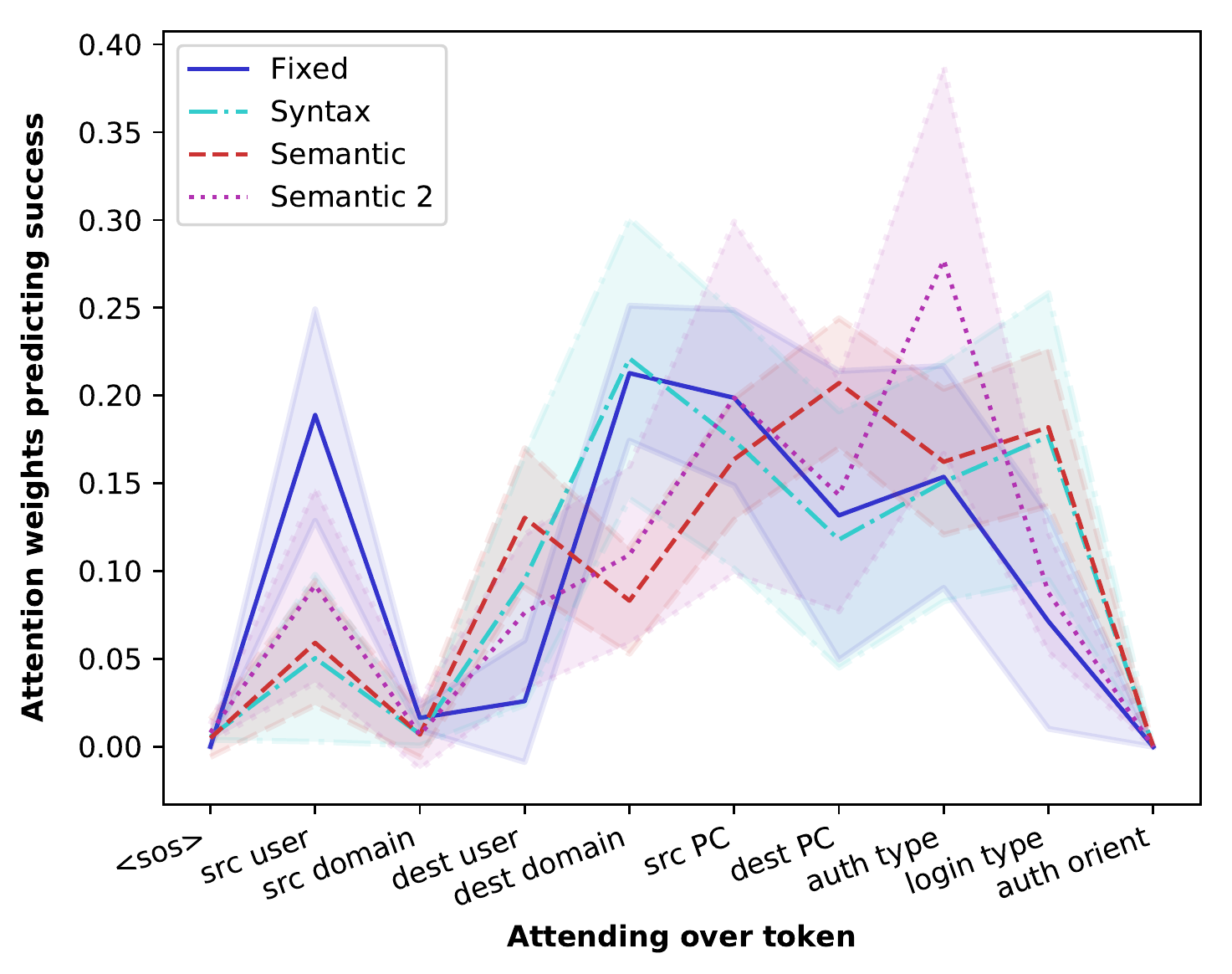}
	\caption{Comparison of attention weights when predicting success/failure token.}
	\label{fig:succfailweights}
\end{figure}

We can gain insight into the global behavior of an attention-equipped LSTM from summary statistics such as the mean and standard deviation of attention weights over the course of a day's predictions. 
Figure \ref{fig:succfailweights} shows the average attention weights for each EM attention model when predicting the last meaningful token (Success/Fail).  Error bars of one standard deviation are shown to illustrate the variability of these weights.  

\begin{figure}
	\includegraphics[width=.5\textwidth]{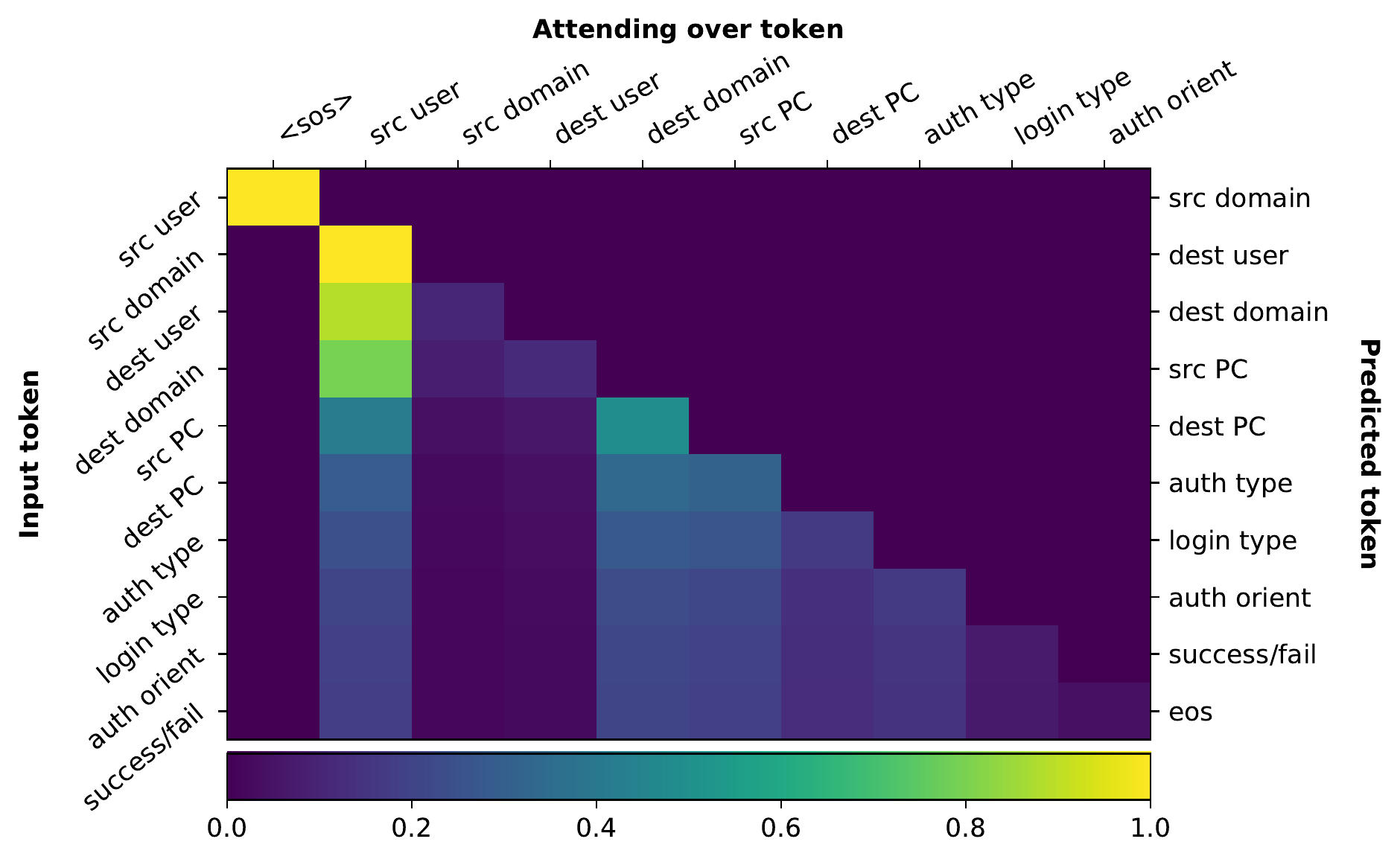}
	\caption{Average Fixed attention weights.} \label{fig:fixedmap}
\end{figure}

Heatmaps of average attention weights for the four EM attention models  proposed in Section \ref{subsec:attention} are provided in Figures \ref{fig:fixedmap}, \ref{fig:syntaxmap}, \ref{fig:semanticmap} and \ref{fig:semantic2map}.
Each time step in a sequence generates its own set of weights, ${\bf d}_{(t)}$, over the previous hidden states. The larger the weight values are the more relevant the associated hidden state is to the current prediction. Note that the first input token is excluded from our figures as it has no previous hidden states to attend over.
\subsubsection{Fixed}
Figure \ref{fig:fixedmap} shows the mean weights for the Fixed attention which has a single fixed query that does not change with the context at the current time step. The source user, destination domain and source PC dominate the weight vectors, suggesting that they are the most important fields to this model. 

\subsubsection{Syntax}
\begin{figure}
	\includegraphics[width=.5\textwidth]{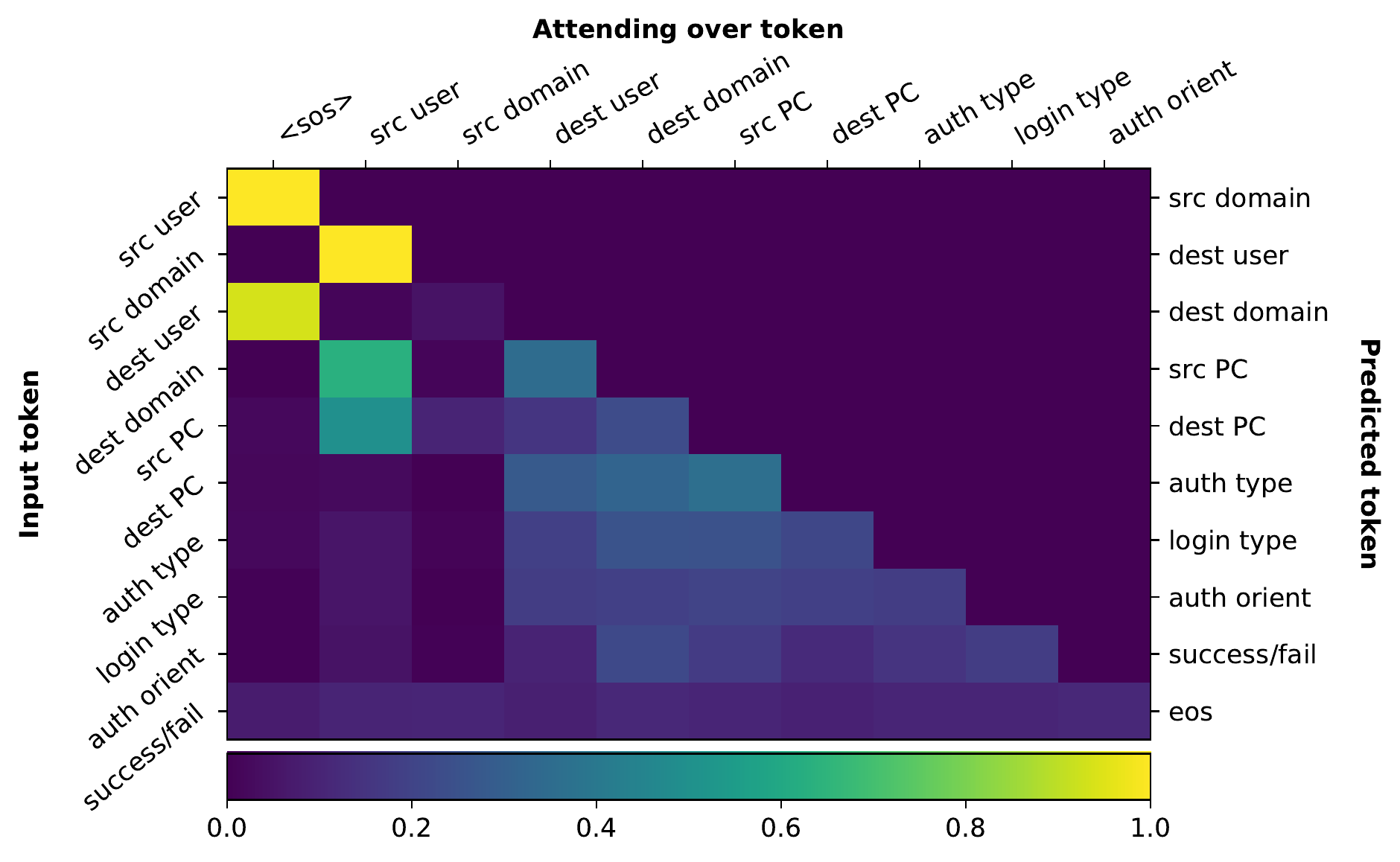}
	\caption{Average Syntax attention weights.} \label{fig:syntaxmap}
\end{figure}
With the syntax model (Figure \ref{fig:syntaxmap}) each time step gets its own set of query weights. This makes sense for word tokenized models that have position dependent syntax. As an example of the model exhibiting intuitive behavior, when predicting the source PC, the model is attending heavily over the source user. 

\subsubsection{Semantic}
\begin{figure}
	\includegraphics[width=.5\textwidth]{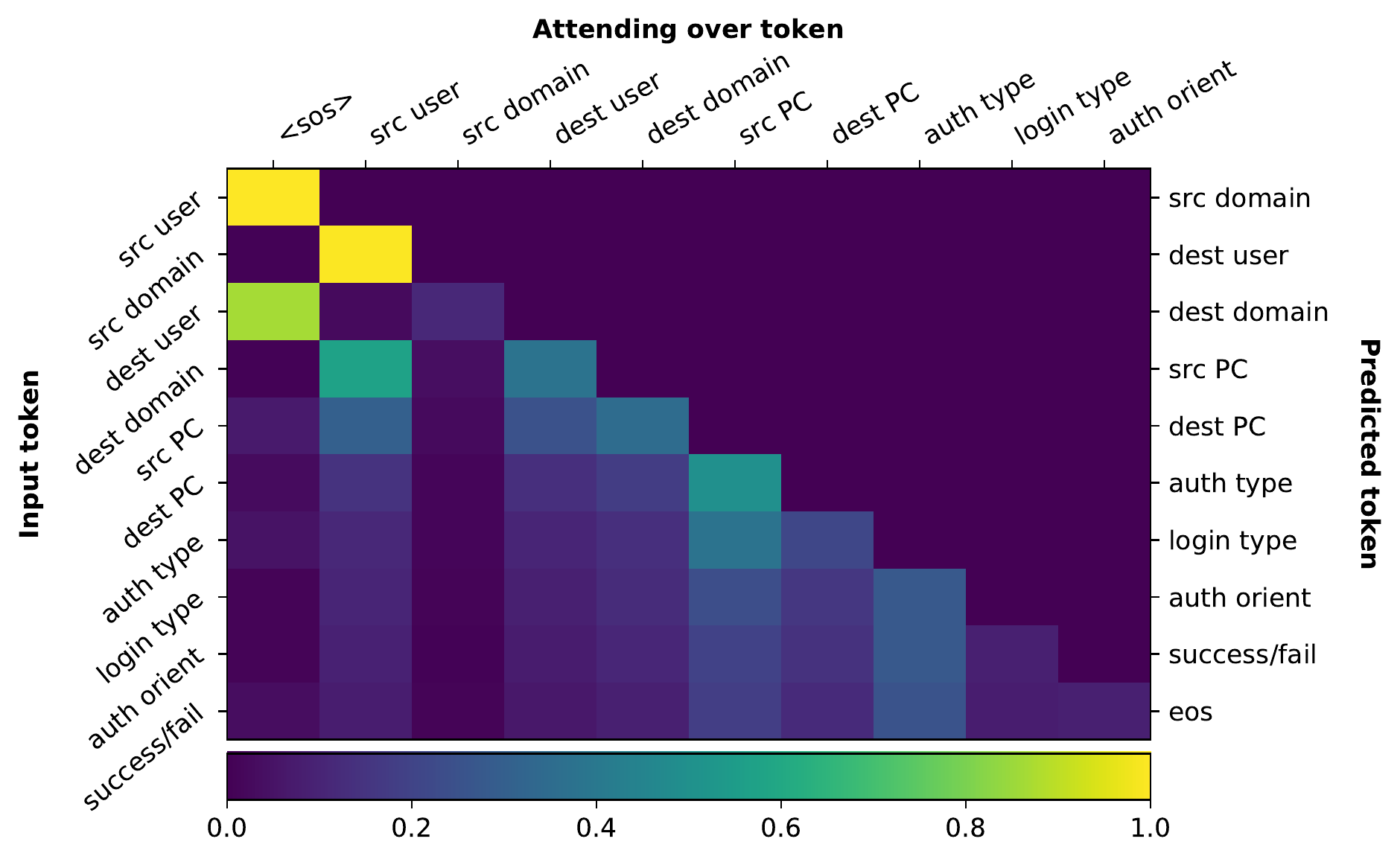}
	\caption{Average Semantic 1 attention weights.} \label{fig:semanticmap}
\end{figure}
\begin{figure}
	\includegraphics[width=.5\textwidth]{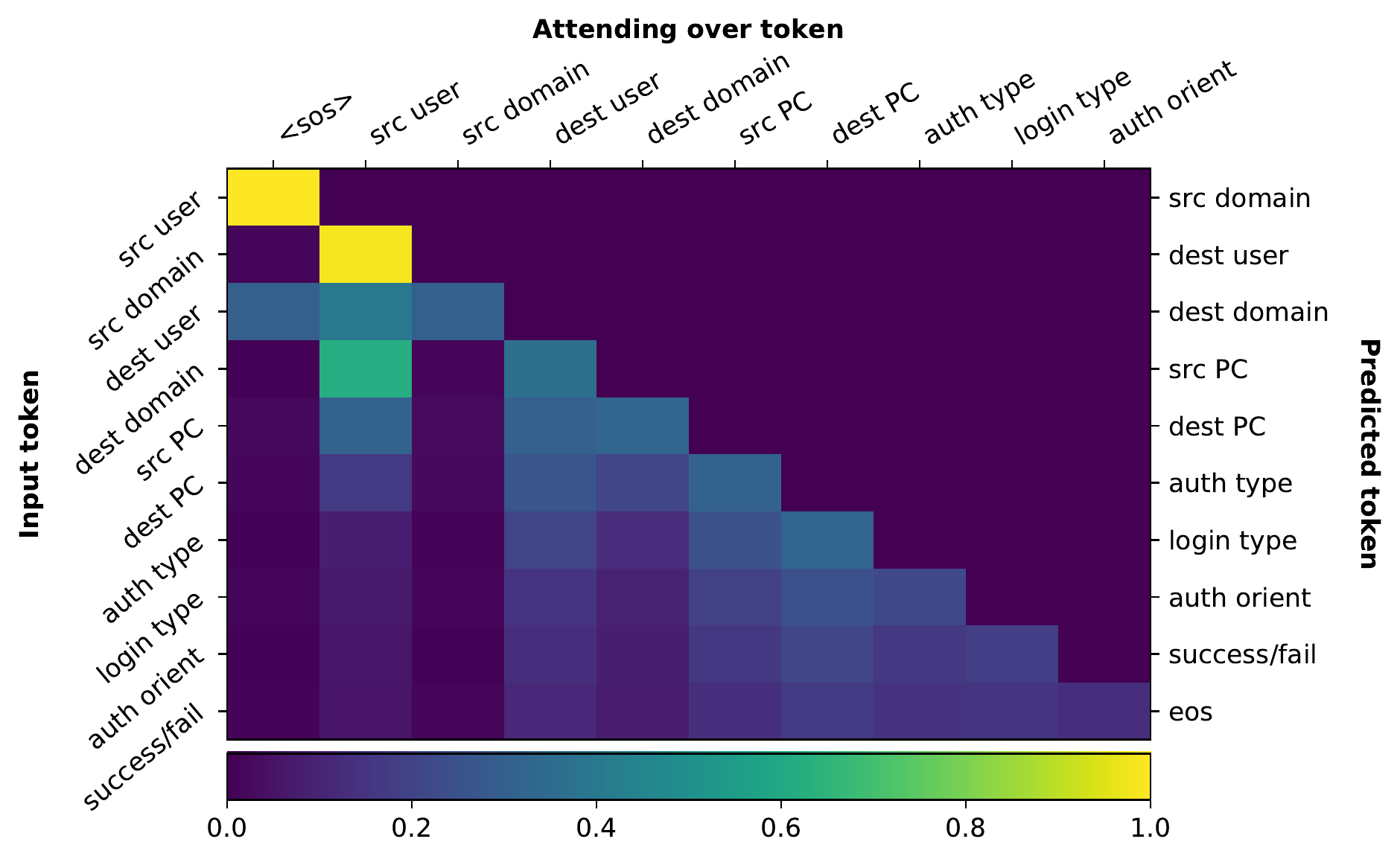}
	\caption{Average Semantic 2 attention weights.} \label{fig:semantic2map}
\end{figure}
While the semantic attention mechanisms do not assume a fixed syntactic structure, Figures \ref{fig:semanticmap} and \ref{fig:semantic2map} show that both semantic attention variants learn reasonable attention strategies on this fixed syntax data. Overall they produce similar attention maps, attending heavily to source user and source PC.  Semantic 1 also attends heavily to authentication type, while Semantic 2 also deems destination user and destination PC to be important.

\subsubsection{Tiered attention models}
For the tiered model with a lower forward-directional LSTM, the attention weights were nearly all 1.0 for the second to last hidden state. This state is making the decision on success/fail, which conceptually makes sense with the goal of top tier LSTM to pass the most relevant information forward for the next event. Conversely, the tiered model with bidirectional LSTM cells attended fully on the very first hidden state. 
As Figure \ref{fig:btier} shows, the backward LSTM ends with the first hidden state. Thus, the bidirectional tiered model is collecting both the final hidden state from the forward LSTM and the backward LSTM as its summary.  This suggests that the shortcut connections attention provides are not needed for this model and task.

\subsection{Case Studies}
\begin{figure}
	\includegraphics[width=0.5\textwidth]{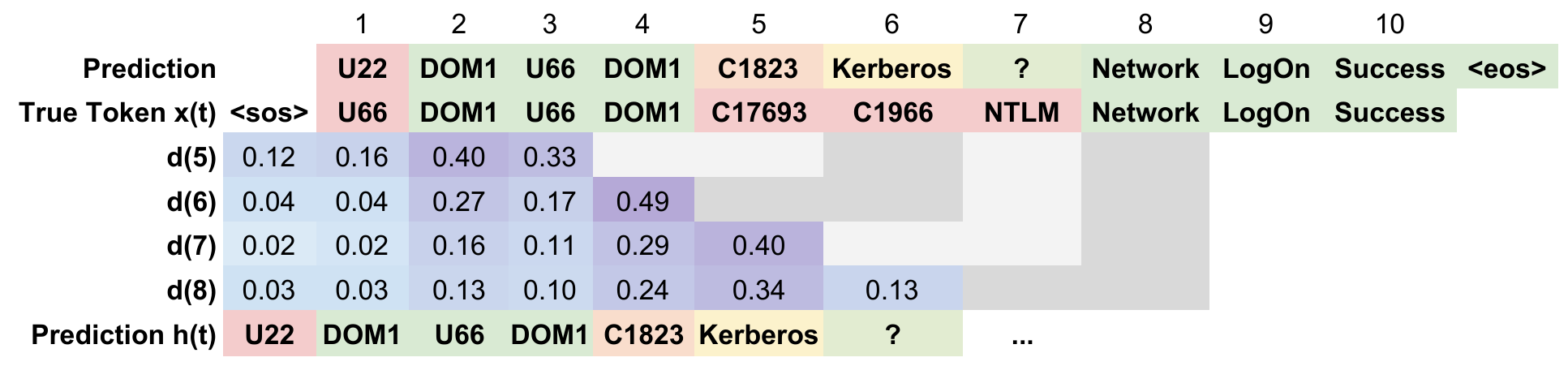}
	\caption{Red team word case study with semantic attention. See Figure \ref{fig:charweights} for description.} \label{fig:wordweights}
\end{figure}

\begin{figure}
	\includegraphics[width=0.5\textwidth]{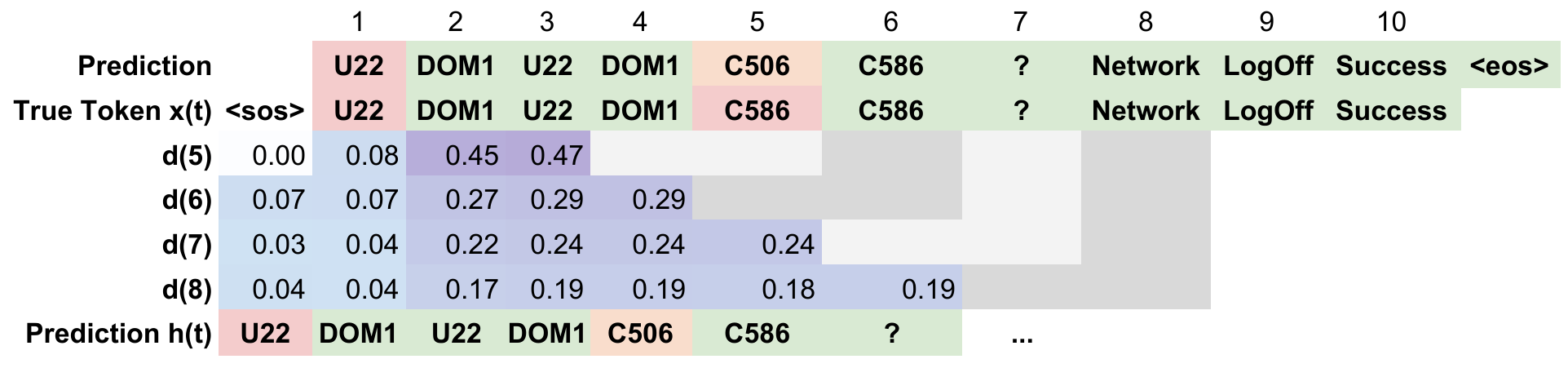}
	\caption{Low anomaly word case study with semantic attention. See Figure \ref{fig:charweights} for description.} \label{fig:wordweights_nonred}
\end{figure}

\begin{figure*}
	\includegraphics[width=1\textwidth]{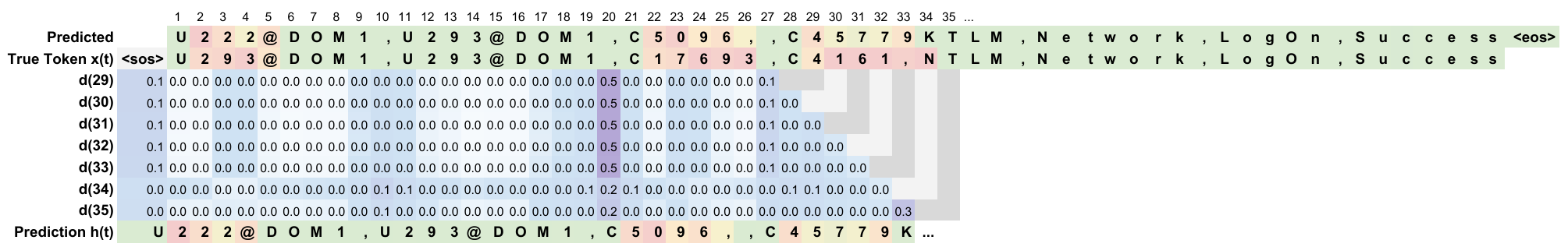}
	\caption{Red team case character study with Semantic attention. Coloring of the true token and predicted token rows is based on the probability of the given character during prediction. Green represents a near 100\% probability while red is near 0\%. Attention weights \textbf{d(t)} correspond to the top row of predictions. For example, when predicting character character 34, K, the model uses attention weights \textbf{d(34)}. We provide a shifted copy of the predicted tokens at the bottom of the figure to align with the hidden states being attended to. Best viewed in color.} \label{fig:charweights}
\end{figure*}

We consider three case studies evaluated using semantic attention models. Figures \ref{fig:wordweights} and \ref{fig:charweights} depict two randomly sampled red events evaluated with word and character semantic attention models, respectively. For contrast, Figure \ref{fig:wordweights_nonred} is a random non-anomalous event evaluated with the semantic word model. Tokens where the predicted and true values diverge are of significant interest as they contribute heavily to the anomaly score. We can disregard the low probabilities when predicting the source user as it is impossible to foresee what user will be associated with a random input sequence. 
\subsubsection{Word Tokenization}
First consider the two word case studies. In both cases the source PC prediction is incorrect with low confidence. 
In the low-anomaly case the model is able to correctly predict the destination PC given the source PC token with very high probability. 
However, the red team event predicted a token associated with a different field for the destination PC. Examining the weights we see that the red team event was attending heavily over the hidden state taking the destination user domain as input and predicting the source user. We note that DOM1 is a very common domain in the LANL dataset and that the attention is likely considering the prediction that will be made from the embedding which relates to the current input token. This misclassification exposes a disadvantage in having a shared vocabulary for each field. Individual vocabularies for each field could improve performance, at the cost of minor feature engineering.

\subsubsection{Character Tokenization}
Finally, we examine model function when processing a character tokenized red team event. When predicting the destination PC characters the hidden state associated with the comma character right before the prediction of the source PC has the largest associated weight. The second largest weight is the comma character right before the destination PC field begins. This may suggest that the model is learning positional information from the comma characters, or that it is accumulating summary vectors of the fields and ``storing them'' in the subsequent delimiter hidden state.
Another point of interest is the attention weight vector d(34). It will substantially impact our anomaly score as our model had almost 100\% confidence that the next character would be `K', while the true token, `N', has near 0\% probability. Again we see a heavy dependence on the delimiter hidden states. 

\section{Conclusions}
In this paper we propose five attention mechanism implementations. 
The fixed and syntactic attention variants can be effective for modeling sequences with a fixed structure while semantic variants are more effective for input sequences that have varying lengths and looser structures. 
While maintaining state-of-the-art performance, the attention mechanisms provide information on both feature importance and relational mapping between features. 
Additionally, architectural insights can be gleaned from the attention applied, which may in the future lead to designing more effective models.
Other future work includes 
evaluating the system on different tasks and domains (e.g. detection of hardware failures from computer logs). 
One could explore additional attention variants; e.g., bidirectional models with attention may lead to further gains in performance. 
Finally, equipping a lower tier model with the ability to attend over upper tier hidden states, may effectively weight the relevance of previous events in a user's log sequence.

\begin{acks}
The research described in this paper is part of the Analysis in Motion Initiative at Pacific Northwest National Laboratory; conducted under the Laboratory Directed Research and Development Program at PNNL, a multi-program national laboratory operated by Battelle for the U.S. Department of Energy. The authors also thank Nvidia  for their donations of Titan X GPU's used in this research.
\end{acks}

\bibliographystyle{ACM-Reference-Format}
\bibliography{attention_interpret}
\end{document}